\documentclass[cameraready]{Interspeech}
\usepackage{graphicx}
\usepackage{pgfplots}
\pgfplotsset{compat=1.18} 

\title{MoVE: Translating Laughter and Tears via Mixture of Vocalization Experts in Speech-to-Speech Translation}

\author[affiliation={1}, orcid=0009-0007-2416-5466, equalcontribution]{Szu-Chi}{Chen}
\author[affiliation={1}, orcid=0009-0005-3127-6004, equalcontribution]{I-Ning}{Tsai}
\author[affiliation={1}, orcid=0009-0007-3994-6433, equalcontribution]{Yi-Cheng}{Lin}
\author[affiliation={2}, orcid=0000-0002-9720-811X, equalcontribution]{Sung-Feng}{Huang}
\author[affiliation={1}, orcid=0000-0002-9654-5747]{Hung-yi}{Lee}

\address{
    $^1$ National Taiwan University, Taipei, Taiwan \\
    $^2$ NVIDIA, Taiwan
}

\email{}

\keywords{speech-to-speech translation, non-verbal vocalizations, mixture of experts, AudioLLMs, expressive speech}

\usepackage{comment}

\begin{document}

\maketitle

\begin{abstract}
Recent Speech-to-Speech Translation (S2ST) systems achieve strong semantic accuracy yet consistently strip away non-verbal vocalizations (NVs), such as laughter and crying that convey pragmatic intent, which severely limits real-world utility. We address this via three contributions. First, we propose a synthesis pipeline for building scalable expressive datasets to overcome the data scarcity limitation. Second, we propose MoVE, a Mixture-of-LoRA-Experts architecture with expressive-specialized adapters and a soft-weighting router that blends experts for capturing hybrid expressive states. Third, we show pretrained AudioLLMs enable striking data efficiency: 30 minutes of curated data is enough for strong performance. On English-Chinese S2ST, while comparing with strong baselines, MoVE reproduces target NVs in 76\% of cases and achieves the highest human-rated naturalness and emotional fidelity among all compared systems, where existing S2ST systems preserve at most 14\% of NVs.

\end{abstract}

\begin{figure*}
    \centering
    \includegraphics[width=1\linewidth]{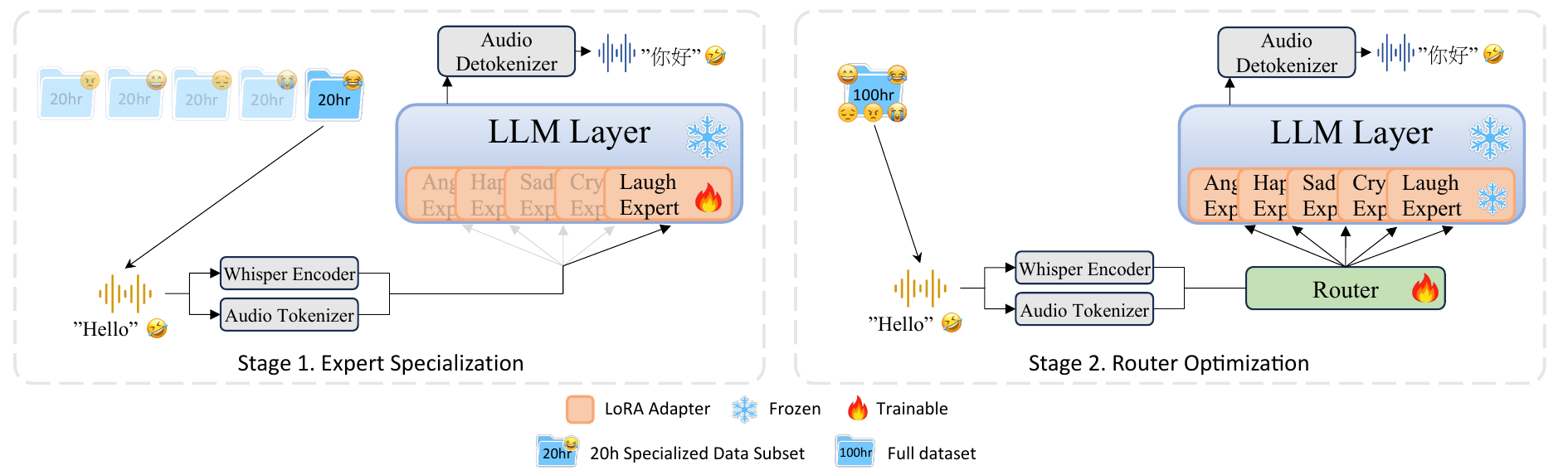}
    \caption{Illustration of MoVE Two-Stage training}
    \label{fig:training_pipeline}
\end{figure*}

\section{Introduction}

Speech-to-Speech Translation (S2ST) represents a sophisticated technology that integrates Automatic Speech Recognition (ASR), Machine Translation (MT), and Text-to-Speech (TTS) synthesis. 
By enabling direct vocal interaction across linguistic boundaries, S2ST transcends the constraints of text-based mediation. 
However, if the translated speech fails to retain the prosody and emotional nuances of the original utterances, it can lead to serious pragmatic biases in cross-language communication~\cite{avila23_interspeech, ROMEROTRILLO201991}. 
For example, a laugh-filled speech translated into a serious tone will lose its humor. 
Similarly, a compliment without the original tone may be misinterpreted as sarcasm. 
Therefore, preserving the expressiveness of the original speech is crucial for significantly improving the naturalness and efficiency of automated S2ST cross-language communication.

While achieving high semantic accuracy, recent S2ST systems~\cite{directs2st, translatotron2, translatotron3, valle-x, seamlessm4t} fundamentally struggle to convey general acoustic expressiveness, much less preserve the specific non-verbal vocalizations (e.g., laughter and crying) that are essential for conveying pragmatic intent.
This gap stems from two fundamental bottlenecks. The most critical bottleneck is the lack of training data. On the data front, high-quality speech corpora containing authentic NVs are extremely scarce and often contaminated by environmental noise, making it difficult for models to disentangle emotional cues from artifacts~\cite{seamlessexpressivelm}. Furthermore, even though there are more and more speech translation~\cite{Seamless, cvss, covost} and non-verbal speech datasets~\cite{nonverbaltts, SMIIP-NV, JVNV}, few of them are suitable for training expressive S2ST, according to their data structure, quality, and quantity. 

Another bottleneck affecting the training of the expressiveness of S2ST lies in the model architecture. In terms of architecture, it is quite difficult to train a custom S2ST model directly on expressive data. 
This is not only because S2ST itself combines three complex tasks: ASR, MT, and TTS, but also because learning cross-linguistic semantic alignment and diverse expressive acoustic preservation further increases the difficulty of training~\cite{directs2st, translatotron2, translatotron3, valle-x, seamlessm4t}. 
AudioLLM~\cite{arora2025on, wu2024audiolanguagemodeling} offers a potential solution because of its large-scale pre-training itself covering tasks such as ASR and TTS, thus providing powerful semantic and acoustic capabilities. However, applying them to the expressive S2ST introduces two additional challenges. First of all, we still cannot be certain how much training data is required to elicit AudioLLM's potential and unlock its S2ST capabilities. Second, how to design an architecture capable of modeling a wide variety of similar or conflicting emotional states accurately, without interference, becomes a crucial issue.


In this paper, we tackle this challenge on two fronts: on the data side, we first propose a scalable expressive dataset synthesis pipeline to overcome training-data limitations, which is further compared with other S2ST datasets. On the model side, we utilize a pre-trained AudioLLM for expressive S2ST fine-tuning, and further discuss the data scale it requires and how we modify the architecture to deal with the multi-emotion modeling problem.
To conclude, the primary contributions of this work are as follows:

\begin{itemize}
    \item \textbf{Pioneering AudioLLM Adaptation for S2ST:} To the best of our knowledge, we are the first to fine-tune general-purpose AudioLLMs for end-to-end S2ST, by effectively eliciting AudioLLMs' powerful potential rather than designing a task-specific architecture.
    \item \textbf{Scalable Expressive Data Synthesis Pipeline:} We introduce an automated generation-selection pipeline tailored for expressive S2ST, encompassing a rich spectrum of affective states alongside specific non-verbal vocalizations like laughter and crying. A generated 1000-hour corpus is released for communities' research purpose\footnote{Dataset and Demo: \url{https://47zzz.github.io/MoVE/}}.
    \item \textbf{MoVE: Mixture of Vocalization Experts with Dynamic Blending:} We propose a framework featuring a learned router that dynamically blends five specialized vocalization adapters, effectively mitigating expressive over-smoothing and significantly improving non-verbal expressive fidelity.
    \item \textbf{Insights on Data Efficiency in AudioLLMs:} Leveraging the robust baseline representations of these foundation models, we demonstrate that fine-tuning via LoRA with as little as \textbf{30 minutes} of curated data achieves 95\% of full-data emotional fidelity without compromising semantic translation accuracy.
\end{itemize}

\section{Methodology}
\subsection{Scalable Expressive Data Synthesis Pipeline}
\label{sec:data_pipeline}

To build a robust foundation for our MoVE training, we propose a scalable pipeline to synthesize an expressive S2ST corpus. Utilizing parallel en-zh text from GigaSpeech and GigaST~\cite{Gigast, gigaspeech}, we generate expressive speech translation pairs through a highly curated emotion-adaptive process:

\textbf{1. Expressive Prompt Curation.}
To prevent the synthesized dataset from degenerating into narrow emotional stereotypes, we establish a high-fidelity acoustic prompt pool. For standard affective states (Happy, Sad, Angry)\footnote{\scriptsize We focus on Happy, Sad, and Angry as they represent the primary extrema of Russell’s affective dimensions~\cite{russell}. Other states, particularly those in the high-valence/low-arousal quadrant, are excluded due to their acoustic proximity to neutral speech, which often leads to classification ambiguity in pure audio. Additionally, a separate 'neutral' class was deemed redundant, as our continuous arousal-based sampling strategy inherently captures neutral-state variations within the broader dataset.}, we aggregate diverse samples across the CREMA-D, MSP-IMPROV, and IEMOCAP datasets~\cite{CREMA-D, MSP-IMPROV, IEMOCAP} to maintain a broad and continuous affective manifold. In contrast, since extreme non-verbal vocalizations (NVs) are often scarce and easily confounded with audio artifacts, we apply more rigorous filtering. Laughter prompts are extracted by applying a laughter detector (confidence $>0.99$)~\cite{Laughter_Seg} to the combined corpora, followed by manual verification. Crying prompts are meticulously sourced from the JVNV dataset~\cite{JVNV}, focusing on human-verified instances of speech-interspersed crying to ensure acoustic viability.

\textbf{2. Emotion-Adaptive Synthesis via Attribute Decoupling.}
We employ IndexTTS2~\cite{indextts2} as our synthesis engine. For standard affective states, the extensive prompt pool allows a single acoustic reference to provide speaker identity and emotional prosody simultaneously. However, the limited availability of curated prompts for extreme NVs poses a challenge to diversity. To mitigate this without sacrificing fidelity, we disentangle identity from expression: the TTS engine is conditioned on a curated NV prompt to reliably induce the target paralinguistics (e.g., crying/laughter), while an independently sampled neutral prompt from the broader corpora defines the speaker identity. This strategy successfully projects rare NV characteristics onto diverse speaker profiles, ensuring comprehensive coverage across five distinct emotional and non-verbal manifolds (Angry, Happy, Sad, Laugh, and Cry).

\textbf{3. Automated Quality Assurance and S2ST Pairing.}
Expressive TTS is prone to hallucinations and text omissions, particularly during NV generation. We apply three sequential filters: (1) silence trimming via librosa, discarding outputs under 0.5 seconds; (2) ASR Word Error Rate (WER) verification using Whisper-small~\cite{whisper} after text normalization, with a lenient threshold of $\leq0.5$ since NV vocalizations yield no consistent lexical transcription and naturally elevate WER; and (3) a pair-level filter that retains a training pair only if both its EN and ZH utterances pass the WER threshold, producing aligned En$\Leftrightarrow$Zh S2ST pairs.


\captionsetup[table]{labelfont=it, textfont=rm}
\setlength{\tabcolsep}{5pt} 
\begin{table*}[t]
\fontsize{8}{9}\selectfont
  \caption{Main expressive S2ST performance. \textbf{Bold} indicates best performance. Values in parentheses represent: (i) bootstrap-estimated standard deviation for \textbf{ASR-BLEU}; (ii) standard error (SE) for \textbf{Aro-Val SIM} ($N=306$); and (iii) SE across participants ($N=5$) for \textbf{Subjective Human Evaluation}. Subjective evaluation (Nat. MOS, Emo. SMOS) and NV Match Accuracy were conducted only for systems included in the architectural ablation and full model comparison; - denotes not evaluated. The symbol * denotes the use of source audio as a one-shot acoustic prompt.}
  \label{tab:main_results}
  \centering
  \begin{tabular}{l c c c c c c}
    \toprule
    & \multicolumn{2}{c}{\textbf{ASR-BLEU} $\uparrow$} & & & & \\
    \cmidrule(lr){2-3}
    \textbf{Model} & \textbf{en$\rightarrow$zh} & \textbf{zh$\rightarrow$en} & \textbf{Aro-Val SIM} $\uparrow$ & \textbf{Nat. MOS} $\uparrow$ & \textbf{Emo. SMOS} $\uparrow$ & \textbf{NV Match (\%)} $\uparrow$ \\
    \midrule
    \multicolumn{7}{l}{\textit{SOTA Foundation \& Commercial Models}} \\
    SeamlessM4T-Large-v2 & 25.8 (1.02)& \textbf{23.6} (1.07)& 0.14 (0.04)& 1.65 (0.20)& 1.47 (0.16)& 2.00 (2.00)\\
    SeamlessExpressive & 23.8 (0.96)& 18.2 (1.02)& 0.45 (0.03)& 1.41 (0.15)& 2.57 (0.24)& 14.00 (9.80)\\
    gpt-4o-audio-preview & 26.3 (1.37)& 19.2 (1.02)& 0.18 (0.04)& 2.87 (0.26)& 1.95 (0.20)& 2.00 (2.00)\\
    Kimi-Audio-7B-Instruct & 25.0 (1.02)& 11.2 (0.91)& 0.11 (0.04)& 3.26 (0.40)& 2.03 (0.17)& 4.00 (2.45)\\
    \midrule
    \multicolumn{7}{l}{\textit{Data \& Architectural Ablations}} \\
    Kimi + LoRA (SeamlessAlign 67h) & 15.7 (0.76)& 12.5 (0.81)& 0.18 (0.04)& - & - & - \\
    Kimi + LoRA (SynStard 100h) & 29.9 (1.12)& 18.4 (0.91)& 0.39 (0.04)& - & - & - \\
    Kimi + LoRA (Ours 50h) & 32.0 (1.12)& 20.1 (1.27)& 0.49 (0.03)& - & - & - \\
    Kimi + LoRA (Ours 100h) & 31.2 (1.12)& 21.2 (0.91)& 0.51 (0.03)& - & - & 26.00 (6.00)\\
    \midrule
    \multicolumn{7}{l}{\textit{Proposed Method}} \\
    \textbf{MoVE}& \textbf{32.5} (1.12)& 21.4 (1.22)& \textbf{0.53} (0.03)& \textbf{3.85} (0.26)& \textbf{3.79} (0.15)& \textbf{76.00} (7.48)\\
    \midrule
    \multicolumn{7}{l}{\color{gray}\textit{Cascaded Oracle}} \\
    \color{gray}Cascaded* & \color{gray}9.7 (0.81)& \color{gray}10.6 (0.86)& \color{gray}0.55 (0.03)& \color{gray}2.61 (0.43)& \color{gray}3.43 (0.25)& \color{gray}26.00 (14.00)\\
    \bottomrule
  \end{tabular}
\end{table*}

\subsection{MoVE Architecture}
\label{ssec:architecture}
We build the MoVE architecture upon a pretrained AudioLLM (Kimi-Audio~\cite{kimi}). To preserve its ASR, TTS and cross-lingual semantic capabilities, we freeze the base parameters and confine expressive adaptation to lightweight LoRA modules~\cite{lora}. Rather than relying on a single module, we implement an X-LoRA-inspired gating strategy~\cite{xlora} to accurately model a wide variety of similar or conflicting emotional states without feature interference.

\textbf{Parallel Expressive Vocalization Experts.}
To prevent conflicting paralinguistic cues from causing mutual interference within a shared parameter space, we inject 5 parallel LoRA adapters across all the transformer layers, each specializing in a distinct acoustic manifold: Happy, Sad, Angry, Laughing, and Crying. Applied to the attention and feed-forward gating projection matrices ($W_q, W_k, W_v, W_o, W_{gate}$)~\cite{xlora}, each adapter operates in an independent low-rank subspace, avoiding feature interference while retaining the base model's capacity for coherent speech generation.

\textbf{Dynamic Soft-Weighting Router.}
To model the inherently hybrid nature of human emotion (e.g., nervous laughter), we eschew hard top-$k$ routing for a dynamic soft-weighting strategy. Let $x \in \mathbb{R}^d$ denote the input hidden state for a speech token, where $d$ is the hidden dimension of the LLM. The MoVE layer output $h(x)$ is computed as the weighted sum of the frozen base weight $W_0$ and the LoRA experts:
$$h(x) = W_0 x + \sum_{i=1}^{N} g_i(x) \cdot (B_i A_i x)$$
where $N=5$ represents the distinct acoustic manifolds, $A_i \in \mathbb{R}^{r \times d}$ and $B_i \in \mathbb{R}^{d \times r}$ are the down-projection and up-projection matrices of the $i$-th LoRA expert with rank $r$, and $g_i(x)$ is a lightweight linear router with a Softmax activation. Operating at the token level, the router assigns a continuous mixture weight to each of the experts for every speech token, enabling fine-grained blending of expert contributions for capturing hybrid expressive states.

\textbf{Expressive Detokenizer.} The pretrained detokenizer, which maps discrete speech tokens to waveforms, fails to faithfully render extreme NVs such as laughter and crying. We fine-tune it on expressive NV speech synthesized with IndexTTS2 using NV-inducing prompts (Section~\ref{sec:data_pipeline}), enabling reliable reconstruction of NV-rich outputs.

\captionsetup[figure]{labelfont=it, textfont=rm}
\subsection{Two-Stage Training Strategy for MoVE}
\label{ssec:training_strategy}
Following the encoder pretraining approach in Kimi-Audio \cite{kimi}, we first fine-tune the Whisper encoder on our expressive corpus to establish a shared acoustic foundation for the subsequent expert modules. MoVE training then proceeds in two stages, as illustrated in Figure~\ref{fig:training_pipeline}:

\textbf{Stage 1: Independent Expert Specialization.}
With the base LLM and aligned Whisper encoder frozen, each of the five LoRA experts is trained independently on its respective expressive subset, ensuring fine-grained specialization within each assigned acoustic manifold while avoiding cross-emotional interference.

\textbf{Stage 2: Dynamic Router Optimization.}
In the final stage, the specialized experts are integrated into the unified MoVE architecture. We optimize the dynamic router on the full dataset while keeping all other parameters fixed. Notably, the router is trained end-to-end via the final language modeling loss rather than explicit labels. This enables the router to autonomously learn the optimal blending of experts based on contextual cues, facilitating the seamless translation of nuanced and hybrid expressive states, effectively bridging the gap between semantic accuracy and emotional fidelity.

\section{Experiments and Analysis}
\subsection{Experimental Setup}

\textbf{Model and Training Dataset Baselines.} 
We compare MoVE against leading end-to-end expressive S2ST systems: Kimi-Audio-7B-Instruct~\cite{kimi}, gpt-4o-audio-preview~\cite{gpt}, SeamlessM4T-Large-v2~\cite{Seamless}, and SeamlessExpressive~\cite{Seamless}. For architectural ablation, we include a single-LoRA baseline fine-tuned on the identical training set; to validate the quality of our data pipeline, we additionally train the same single-LoRA model on SynStard-1000 (randomly sample 100h)~\cite{SynStard} and SeamlessAlignExpressive (67h)~\cite{Seamless} as data-quality comparisons. For reference, we include a cascaded oracle (source audio as a one-shot acoustic prompt) as an upper bound with pipeline (Whisper-large-v3~\cite{whisper} $\rightarrow$ NLLB-200~\cite{NLLB} $\rightarrow$ CosyVoice3-0.5B~\cite{cosyvoice}), though it is not directly comparable given its fundamentally different system paradigm.

\noindent\textbf{Test Sets and Evaluation Metrics.} We decouple the evaluation into three logical dimensions: semantic preservation, objective emotional fidelity, and subjective human perception.
\textit{1) Semantic Translation:} We evaluate linguistic accuracy using ASR-BLEU~\cite{BLEU} on 1000 English-Chinese pairs sampled from CVSS-T~\cite{cvss, covost}, including English-to-Chinese and Chinese-to-English translation.
\textit{2) Objective Emotional Fidelity:} We curate a test set from the NonverbalTTS corpus~\cite{nonverbaltts} (strictly filtering out segments lacking linguistic content) encompassing Happy, Sad, Neutral, and Angry states. Objective fidelity is measured via Arousal-Valence Similarity (\textit{Aro-Val SIM}) --- the cosine similarity between source and generated arousal-valence score~\cite{aro-val, laugh_cry}.
\textit{3) Subjective Human Evaluation:} We uniformly sample 30 representative utterances across six evaluation categories (Happy, Sad, Angry, Laughing, Crying, Neutral). 
To ensure evaluation rigor while managing the cognitive load of the annotator, we adopt a dual-paradigm testing strategy. 
For comparisons against external and cascaded baselines, proficient bilingual evaluators conduct a multi-stimulus test to evaluate \textit{Naturalness MOS} (Mean Opinion Score) and \textit{Emotion SMOS} (Similarity Mean Opinion Score)~\cite{wester2016analysis}, rating the generated speech on a standard 1-to-5 scale. Furthermore, we compared our MoVE framework against the internal single-LoRA baseline using a pairwise A/B preference test as the model architecture ablation, where evaluators strictly judge superior emotional expressiveness (MoVE vs. Single-LoRA, including a ``Tie'' option). Finally, across all evaluated models, evaluators assess \textit{NV Match Accuracy} for the two extreme NV categories, recording a ``hit'' only if the explicitly perceived NVs exactly align with the source. The complete evaluation interface, instructions, anchor calibration, and per-phase procedure are detailed in Appendix~\ref{sec:appendix_eval}.

\noindent\textbf{Implementation Details.} LoRA experts are configured with rank $r=256$ and scaling factor $\alpha=256$. The model is optimized using AdamW ($\beta_2=0.95$) with a learning rate of $1 \times 10^{-5}$. To ensure deep feature disentanglement, Stage 1 (expert specialization) is trained for 2 epochs, while Stage 2 converges effectively within 1 epoch.

\subsection{Main Expressive S2ST Performance}

Table \ref{tab:main_results} shows our main experiment results. We compare with different model baselines that do not require further training and also compare different training datasets with our single-LoRA baseline model, where we sampled all the datasets to a similar size between 50h to 100h. Lastly, we show our MoVE performance and also compare with cascaded pipeline, while it mostly serves as a reference given its fundamentally different system paradigm. As shown in Table \ref{tab:main_results}, MoVE establishes a new state-of-the-art in expressive S2ST. It achieves a new SOTA ASR-BLEU on en$\rightarrow$zh (32.5), and is highly competitive on zh$\rightarrow$en (21.4), trailing SeamlessM4T-Large-v2 by 2.2 BLEU on zh→en (21.4 vs. 23.6), a gap attributable to MoVE's optimization priority on expressive fidelity over pure semantic accuracy. This success is attributed to two key contributions: our proposed expressive dataset synthesis pipeline and our MoVE architectural design.

\textbf{Training Dataset Comparison.}
To strictly isolate and validate the quality of our proposed expressive corpus (Section 3.1), we restrict all evaluations in this part to the objective metrics. We compare models fine-tuned with a single LoRA on our dataset (sampled at 50h and 100h) against those trained on the randomly sampled 100h SynStard subset and the 67h SeamlessAlignExpressive corpus. Results indicate that even the 50h subset of our data overwhelmingly surpasses both external baselines in ASR-BLEU and Aro-Val SIM. Furthermore, the 100h version of our dataset consistently outperforms the 100h SynStard subset under the identical single-LoRA architecture, confirming that the quality advantage stems from our data pipeline rather than data volume or model capacity.

\textbf{MoVE Architectural Comparison.}
Comparing MoVE with our single-LoRA baseline (both trained with the same 100h data), MoVE surpasses throughout all the evaluation metrics, proving that reallocating model capacity to master expressive manifolds does not compromise semantic accuracy.
Subjectively, MoVE secures the highest Emotion SMOS and Naturalness MOS. In A/B preference tests (see Table \ref{tab:ab_test}), it overwhelmingly outperforms the single-LoRA baseline. Furthermore, MoVE dominates in NV Match Accuracy, successfully synthesizing extreme paralinguistics (e.g., laughter, crying) where single-LoRA baseline typically failed to preserve, while it is hard to exhibit such observation through the given objective score.
Objectively, MoVE's Aro-Val SIM trails the cascaded pipeline by a negligible margin, while outstandingly win over the cascaded pipeline in ASR-BLEU. Although this comparison is architecturally asymmetrical\footnote{\scriptsize As the cascaded system utilizes one-shot source audio prompting instead of pure zero-shot inference.}, MoVE achieves comparable objective fidelity while significantly surpassing the cascaded pipeline across all human-perceived metrics, underscoring the dynamic router's superior affective rendering.

\begin{table}[h]
\fontsize{8}{9}\selectfont
  \centering
  \begin{tabular}{l c}
    \toprule
    \textbf{Preference} & \textbf{Win Rate (\%)} \\
    \midrule
    \textbf{MoVE} & \textbf{60.00} \\
    Tie & 22.67 \\
    Single-LoRA (Ours 100h) & 17.33 \\
    \bottomrule
  \end{tabular}
  
  \vspace{8pt} 
  
  \caption{A/B Test: Overall Preference (Win Rate \%) between MoVE and the best Single-LoRA baseline.}
  \label{tab:ab_test}
\end{table}

\subsection{Analysis on Data Scale Efficacy}

\begin{figure}[t]
\centering
\begin{tikzpicture}
\pgfplotsset{set layers}

\begin{axis}[
    width=.8\linewidth,
    height=4.5cm,
    xlabel={Training Set Scale (Hours)},
    symbolic x coords={0, 0.1, 0.5, 1, 5, 10, 50, 100, 500, 1000},
    xtick=data,
    grid=major,
    grid style={dashed, gray!20},
    ylabel={ASR-BLEU Score},
    ylabel style={blue},
    yticklabel style={blue},
    axis y line*=left,
    ymin=-1.5, ymax=31, 
    legend style={
        at={(0.5,-0.25)},
        anchor=north,
        legend columns=2, 
        font=\tiny,
        draw=none, 
        /tikz/every even column/.append style={column sep=10pt}
    },
    tick label style={font=\tiny},
    label style={font=\scriptsize},
]

    \addplot[blue, mark=circle, thick, error bars/.cd, y dir=both, y explicit]
    coordinates {
        (0, 0) (0.1, 19.08)+-(0,0.65) (0.5, 23.99)+-(0,0.89) (1, 24.44)+-(0,0.70)
        (5, 25.04)+-(0,0.79) (10, 25.84)+-(0,0.85) (50, 25.94)+-(0,0.85)
        (100, 26.11)+-(0,0.72) (500, 26.73)+-(0,0.77) (1000, 27.0)+-(0,0.79)
    };
    \addlegendentry{BLEU Score: Kimi+LoRA}

    \addlegendimage{red, mark=circle, thick}
    \addlegendentry{Aro-Val SIM: Kimi+LoRA}

    \addplot[blue, dashed, mark=circle, thick]
    coordinates { (0.1, 0.05) (5, 0.05) (50, 0.05) (1000, 0.05) };
    \addlegendentry{BLEU Score: (Random Init)}

    \addlegendimage{red, dashed, mark=circle, thick}
    \addlegendentry{Aro-Val SIM: (Random Init)}

\end{axis}

\begin{axis}[
    width=.8\linewidth,
    height=4.5cm,
    axis x line=none,
    axis y line*=right,
    ylabel={Aro-Val SIM Score},
    ylabel style={red},
    yticklabel style={red},
    symbolic x coords={0, 0.1, 0.5, 1, 5, 10, 50, 100, 500, 1000},
    ymin=-0.03, ymax=0.62,
    legend style={
        font=\tiny,
    },
    tick label style={font=\tiny},
    label style={font=\scriptsize},
]
    \addplot[red, mark=circle, thick, error bars/.cd, y dir=both, y explicit]
    coordinates {
        (0, 0.0539)+-(0,0.04) (0.1, 0.3662)+-(0,0.03) (0.5, 0.4538)+-(0,0.03)
        (1, 0.4671)+-(0,0.03) (5, 0.4917)+-(0,0.03) (10, 0.5069)+-(0,0.03)
        (50, 0.4907)+-(0,0.03) (100, 0.506)+-(0,0.03) (500, 0.4625)+-(0,0.03) (1000, 0.4802)+-(0,0.03)
    };

    \addplot[red, dashed, mark=circle, thick, error bars/.cd, y dir=both, y explicit]
    coordinates {
        (0.1, 0.0591)+-(0,0.04) (5, 0.0739)+-(0,0.04) (50, 0.2052)+-(0,0.04) (1000, 0.206)+-(0,0.04)
    };

\end{axis}
\end{tikzpicture}
\caption{Data efficacy and scaling behaviors across varying training sizes. The left axis corresponds to ASR-BLEU and the right axis to Aro-Val SIM. ASR-BLEU is the average of en-zh and zh-en. Error bars represent standard errors.}
\label{fig:combined_scaling}
\end{figure}
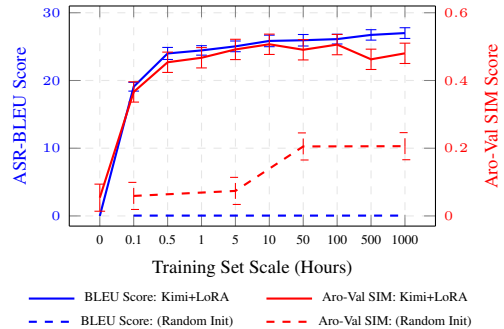

To investigate data efficiency, we scale the training subset across 0, 0.1, 0.5, 1, 5, 10, 50, 100, 500, and 1000 hours, where 0 hours represents the original base Kimi model without additional training. Since we're using objective evaluation for the data efficiency ablation, which our single-LoRA baseline performs on par with MoVE, we eventually choose the single-LoRA architecture for simplicity. As illustrated in Figure~\ref{fig:combined_scaling}, the learning curves for both semantic and emotional fidelity exhibit a striking phenomenon: while performance drops sharply at 0.1 hours, it saturates rapidly, maintaining a robust plateau from 0.5 hours all the way to 1000 hours. To determine the origin of this extreme data efficiency, we conduct an ablation study by re-initializing the weights of the base AudioLLM from scratch and training it on varying subsets (0.1h, 0.5h, 50h, 1000h). Under this condition, the model completely fails to converge, collapsing into unintelligible babble across all data scales (represented by the blue dashed lines in Figure \ref{fig:combined_scaling}.)
This stark contrast reveals a key insight: the data efficiency we observe comes not from LoRA's adaptive capacity itself, but from the acoustic and semantic knowledge already stored in the AudioLLM's pre-trained weights, which LoRA activates rather than creates.

\captionsetup[figure]{labelfont=it, textfont=rm}

\subsection{Router Behavior and Disentanglement}

\begin{figure}
    \centering
    \includegraphics[width=0.8\linewidth]{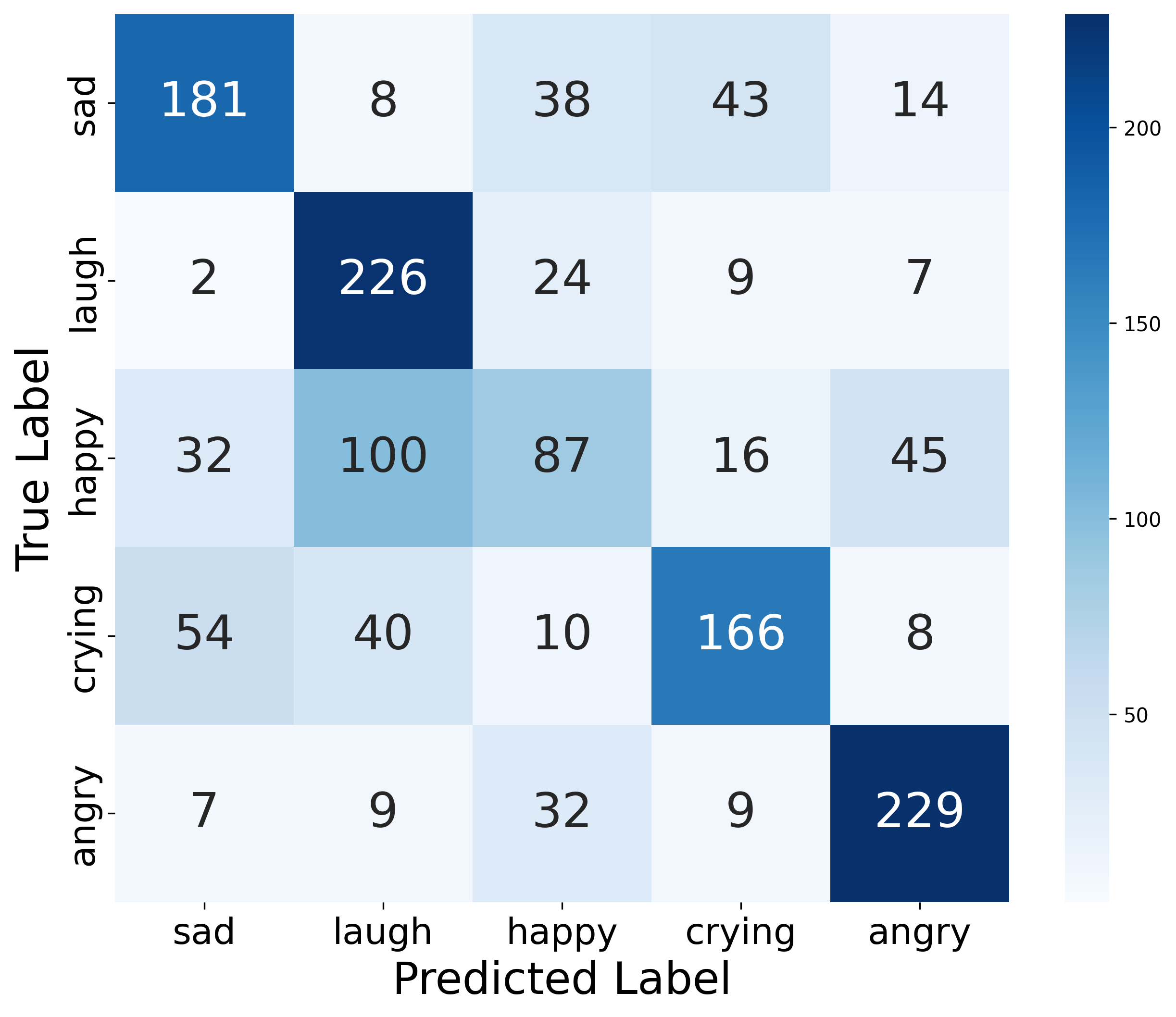}
    \caption{Confusion matrix of router behavior.}
    \label{fig:confusion_matrix}
\end{figure}

To interpret the internal routing mechanism of our MoVE architecture, we visualize the expert activation patterns on an unseen development set. Figure \ref{fig:confusion_matrix} presents a confusion matrix that evaluates the alignment between the predefined expressive categories (Angry, Happy, Sad, Laugh, Cry) and the dominant expert selection of the router. 

As described in Section~\ref{ssec:training_strategy}, the router is optimized end-to-end only through the loss of final language modeling, without ever being exposed to discrete emotion labels. However, benefiting from the supervised specialization of the underlying experts, the router eventually achieves an overall alignment accuracy of 63.68\%.  Achieving such a high alignment accuracy under these unconstrained conditions strongly demonstrates that the router autonomously learns to disentangle affective states by capturing underlying latent linguistic and acoustic cues. 

Furthermore, the off-diagonal distributions in the confusion matrix (e.g., routing overlaps between ``Sad'' and ``Cry'', or ``Happy'' and ``Laugh'') are not arbitrary errors. Rather, they physically reflect the nuanced and hybrid nature of natural human expressiveness. This unsupervised soft-weighting capability validates the core motivation of our MoE design, proving it can dynamically modulate expressive blends far beyond rigid, rule-based classification.

\section{Conclusions}

This paper addresses the expressive gap in S2ST. We proposed a scalable, expressive data curation pipeline for training and demonstrated its superiority over other datasets. By leveraging the robust priors of pre-trained AudioLLMs, our MoVE achieves state-of-the-art fidelity in transferring emotions and NVs with incredible data efficiency: as little as 30 minutes of curated data can "unlock" deep expressive capabilities. Although not covering all emotions and NV categories, our method has the potential to be extended to a more real-world scenario. Our curated dataset will also be released after camera-ready.


\section{Acknowledgments}
This work was supported by the Ministry of Education (MOE) of Taiwan under the Taiwan Centers of Excellence in Artificial Intelligence project, through the NTU Artificial Intelligence Center of Research Excellence (NTU AI-CoRE). Computing resources were provided by the National Center for High-Performance Computing, National Institutes of Applied Research (NIAR), Taiwan.

\section{Generative AI Use Disclosure}
We employed Gemini for grammatical paraphrasing and language polishing to improve the manuscript's clarity.

\bibliographystyle{IEEEtran}
\bibliography{mybib}

\appendix
\section{Subjective Human Evaluation Protocol}
\label{sec:appendix_eval}

We complement Section~4.1 with a brief account of the in-house bilingual evaluation platform used for all subjective scores. Five proficient English--Chinese bilingual evaluators ($N=5$) rate the 30-utterance test set (six categories $\times$ five utterances). For every evaluator, both the set order and the model order within each set are independently randomized via a per-subject seed.

\begin{figure}[h]
    \centering
    \includegraphics[width=\linewidth]{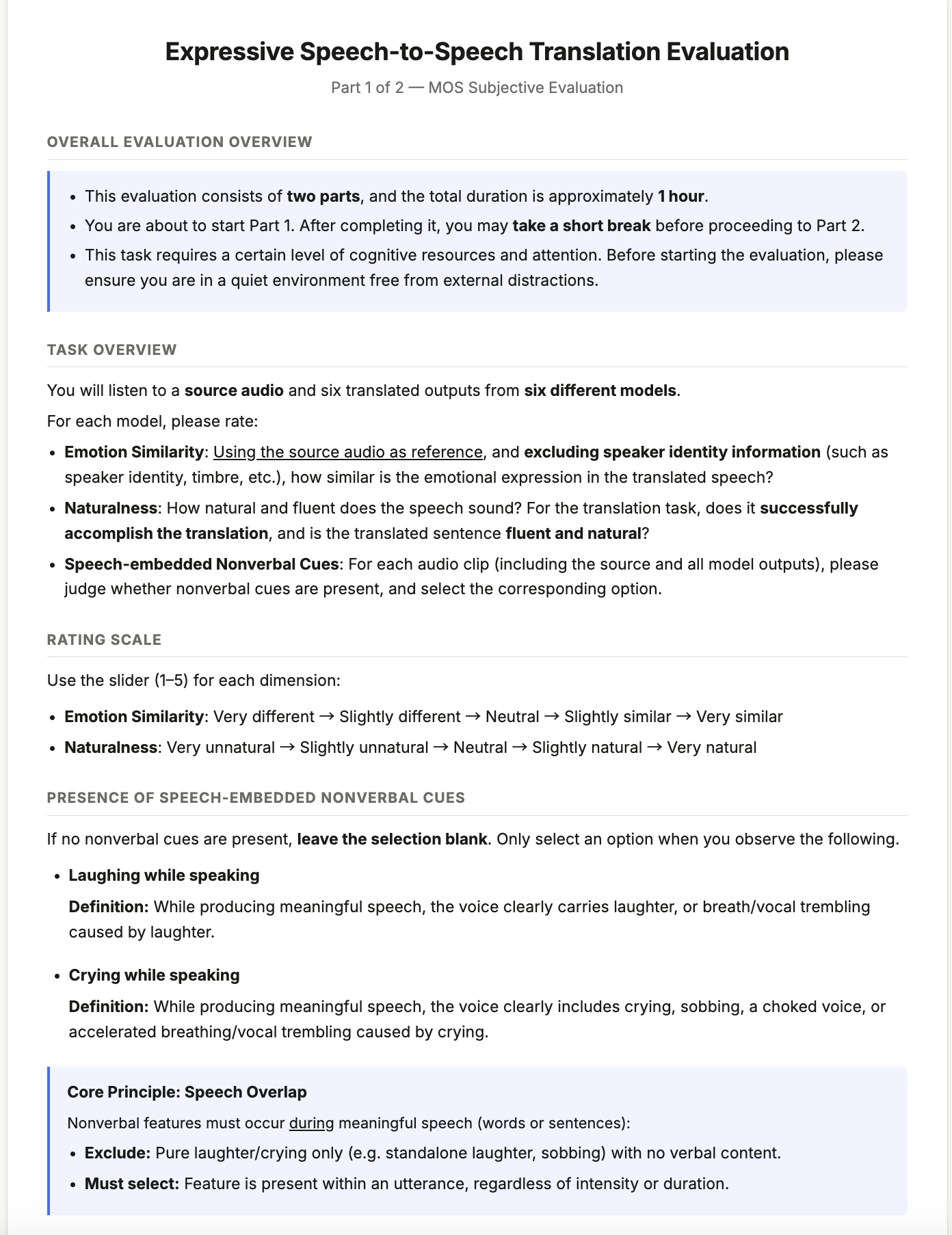}
    \caption{Bilingual instruction page shown before Phase 1 (MOS evaluation): defines the Emotion Similarity and Naturalness rating scales (1–5), the two NV categories (Laughing / Crying while speaking), and the Speech Overlap principle requiring NVs to co-occur with linguistic content.}
    \label{fig:instruction}
\end{figure}

\textbf{Instructions and anchor calibration} (Figure~\ref{fig:instruction}). Before each phase, evaluators read a bilingual instruction page that defines the rating scale, the two NV categories (\textit{Laughing} / \textit{Crying while speaking}), and a \textbf{Speech Overlap} principle requiring NVs to co-occur with linguistic content. 

\begin{figure*}[h]
    \centering
    \includegraphics[width=\linewidth]{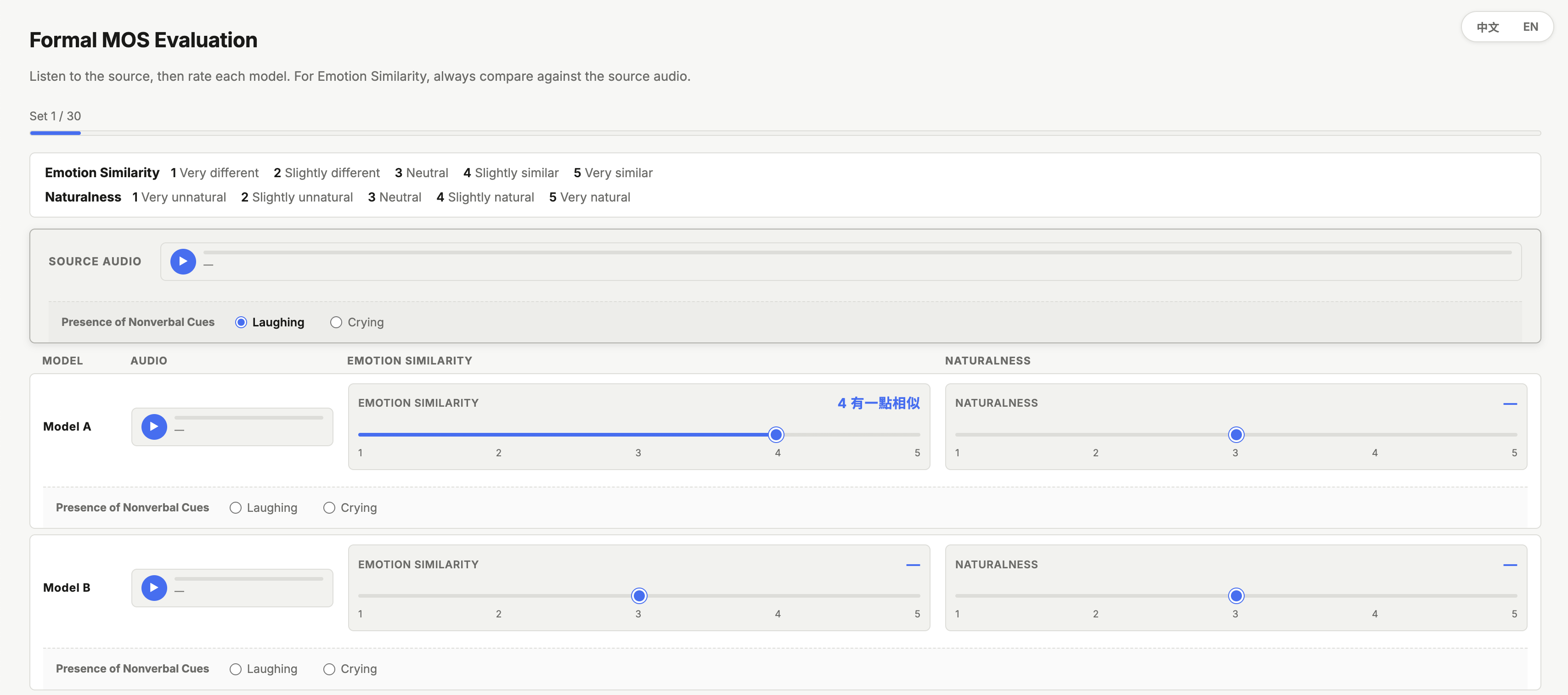}
    \caption{Phase 1 MOS multi-stimulus interface: evaluators listen to the source audio (top, used as the emotional reference) and rate each of the six anonymized model outputs on Emotion Similarity and Naturalness (1–5 sliders). An NV selector (Laughing / Crying) is provided for each clip to record perceived non-verbal vocalizations.}
    \label{fig:mos_eval}
\end{figure*}

\textbf{Phase 1: MOS multi-stimulus test} (Figure~\ref{fig:mos_eval}). Each screen shows the source audio above six anonymized, randomly labeled model outputs (MoVE, cascaded, SeamlessExpressive, Kimi-Audio, SeamlessM4T, gpt-4o-audio). For each output, evaluators score \textit{Emotion Similarity} and \textit{Naturalness} on independent 1--5 sliders, and additionally mark any perceived NV cue. Source-side NV annotations are also collected and serve as the per-utterance ground truth for NV Match Accuracy.

\begin{figure*}[h]
    \centering
    \includegraphics[width=\linewidth]{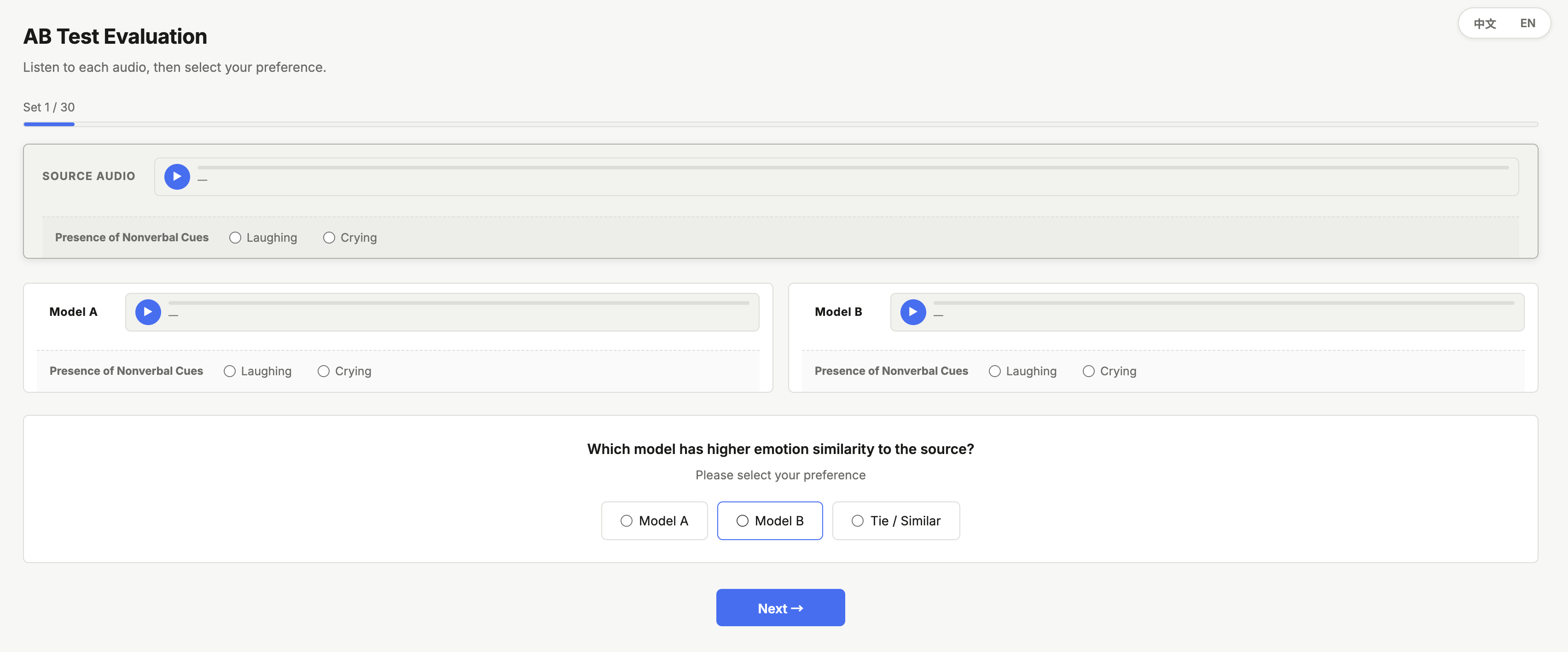}
    \caption{Phase 2 pairwise A/B interface for the MoVE vs. single-LoRA architectural ablation: evaluators judge which model has higher emotional expressiveness (Model A, Model B, or Tie / Similar). Model identity and side assignment are randomized per utterance. An NV selector is provided for cross-validation against the source.}
    \label{fig:ab_eval}
\end{figure*}

\textbf{Phase 2: A/B pairwise preference test} (Figure~\ref{fig:ab_eval}). For the architectural ablation, MoVE and the single-LoRA baseline are presented side by side with per-set randomized A/B labeling. Evaluators choose \textit{Model A}, \textit{Model B}, or \textit{Tie / Similar}, with \textit{Tie} restricted to cases that are perceptually indistinguishable. Per-clip NV selectors are again provided for cross-validation against the source.

\textbf{NV Match Accuracy.} For each utterance the source-side annotation defines the ground-truth NV label; a model scores a \textit{hit} only when its perceived NV label exactly matches the source (otherwise a miss). Per-system accuracies are averaged across the laughing and crying subsets, and standard errors over the five evaluators are reported in Table~\ref{tab:main_results}.

\end{document}